\documentclass[runningheads]{llncs}
\usepackage{graphicx}
\usepackage{makecell}
\usepackage{hyphenat}
\usepackage{hyperref}
\hypersetup{
    colorlinks=true,
    linkcolor=blue,
    filecolor=magenta,      
    urlcolor=cyan,
    pdftitle={Overleaf Example},
    pdfpagemode=FullScreen,
    citecolor=blue,
    }

\usepackage[table, dvipsnames]{xcolor}
\usepackage{cite}
\usepackage{booktabs}
\usepackage{multirow}
\usepackage{nicematrix}
\usepackage[T1]{fontenc}

\renewcommand{\arraystretch}{2}
\arrayrulecolor[HTML]{000000}

\definecolor{DatasetNameColumn_bc}{HTML}{FF9523}
\colorlet{DatasetNameColumn_title_color}{DatasetNameColumn_bc!125!black!220}
\colorlet{DatasetNameColumn_color}{DatasetNameColumn_title_color!20}
\colorlet{DatasetNameColumn_color_light}{DatasetNameColumn_title_color!5}

\definecolor{OtherColumns_bc}{HTML}{1E3143}
\colorlet{OtherColumns_title_color}{OtherColumns_bc!95}

\colorlet{Gray_OddRows}{gray!15}
\colorlet{Gray_EvenRows}{gray!40}

\definecolor{ResNet_bc}{HTML}{1E3143}
\colorlet{ResNet_color}{ResNet_bc!95}

\definecolor{EndoViT_bc}{HTML}{FF9523}
\colorlet{EndoViT_color}{EndoViT_bc!125!black!220}

\definecolor{MAE_NoPretraining_bc}{HTML}{ce99ff}
\colorlet{MAE_NoPretraining_color}{MAE_NoPretraining_bc!50!black}

\definecolor{MAE_ImageNet_bc}{HTML}{38bfa7}
\colorlet{MAE_ImageNet_color}{MAE_ImageNet_bc!80!black}

\definecolor{Stage1_bc}{HTML}{0278b8}
\colorlet{Stage1_color}{Stage1_bc!35}
\colorlet{Stage1_light_color}{Stage1_bc!15}
\colorlet{Stage1_lighter_color}{Stage1_bc!5}

\definecolor{Stage2_bc}{HTML}{F5D76E}
\colorlet{Stage2_color}{Stage2_bc!45}
\colorlet{Stage2_light_color}{Stage2_bc!30}
\colorlet{Stage2_lighter_color}{Stage2_bc!10}

\definecolor{VideosOnly_bc}{HTML}{F5D76E} 
\colorlet{2VideosOnly_color}{VideosOnly_bc!20}
\colorlet{4VideosOnly_color}{VideosOnly_bc!40}
\colorlet{8VideosOnly_color}{VideosOnly_bc!20}

\begin{document}
\title{Whether and When does Endoscopy \\Domain Pretraining Make Sense?}
\author{Dominik Batić\inst{}\thanks{All authors share first authorship.}, 
Felix Holm\inst{\star}, 
Ege Özsoy\inst{\star}, 
Tobias Czempiel\inst{},
Nassir Navab\inst{}}
%
\authorrunning{D. Batić, F. Holm, E. Özsoy, T. Czempiel, N. Navab}
\institute{Computer Aided Medical Procedures, Technical University Munich, Garching, Germany}

\maketitle
\begin{abstract}
Automated endoscopy video analysis is a challenging task in medical computer vision, with the primary objective of assisting surgeons during procedures. The difficulty arises from the complexity of surgical scenes and the lack of a sufficient amount of annotated data. In recent years, large-scale pretraining has shown great success in natural language processing and computer vision communities. These approaches reduce the need for annotated data, which is always a concern in the medical domain. However, most works on endoscopic video understanding use models pretrained on natural images, creating a domain gap between pretraining and finetuning. In this work, we investigate the need for endoscopy domain-specific pretraining based on downstream objectives. To this end, we first collect Endo700k, the largest publicly available corpus of endoscopic images, extracted from nine public Minimally Invasive Surgery (MIS) datasets. Endo700k comprises more than 700,000 unannotated raw images. Next, we introduce EndoViT, an endoscopy pretrained Vision Transformer (ViT). Through ablations, we demonstrate that domain-specific pretraining is particularly beneficial for more complex downstream tasks, such as Action Triplet Detection, and less effective and even unnecessary for simpler tasks, such as Surgical Phase Recognition. We will release both our code and pretrained models upon acceptance to facilitate further research in this direction.

\end{abstract}

\section{Introduction}
Minimally Invasive Surgery (MIS) is quickly becoming one of the most common styles of surgical procedures in the world~\cite{rendezvous}. In contrast to open surgery, MIS lowers the chance of infection and speeds up the recovery rate. As MIS procedures use endoscopic cameras, it has become possible to analyze large amounts of video data, leading to the development of surgical assistance systems. These systems can detect errors and provide decision support to improve patient outcomes~\cite{SurgicalDataScience}. Additionally, recorded surgical procedures are being categorized, providing valuable insights to surgeons, enabling them to learn and improve their techniques from hours of surgical recordings~\cite{SurgicalActions160}. To achieve these goals, the community has already thoroughly investigated the task of Surgical Phase Recognition \cite{twinanda2016endonet, tecno} as well as successfully managed to detect and localize surgical instruments \cite{jha2021kvasir}. Today, more challenging tasks are being explored, such as the newly introduced Action Triplet Detection \cite{rendezvous}. It requires not only detecting surgical instruments, actions and anatomies but also determining the relationship between them. Other works focus on segmentation of tools and tissues\cite{chen2021transunet}, as well as multi-level learning, combining several tasks at once\cite{PSI_AVA}. 

 On the other hand, in recent years, the transformer~\cite{vaswani2017attention} architecture, based on self-attention mechanisms, has had a tremendous impact on deep learning. The transformer models' success can be attributed to the introduction of self-supervised pretraining methods, such as Masked Language Modeling. The idea is straightforward: a percentage of input words are randomly masked out, and the model is tasked with predicting the missing input. Despite its simplicity, it presents a challenging self-supervised task. This approach has led to a paradigm shift in which a transformer network is first pretrained on large amounts of unlabelled data in order to create a model with a general understanding of the underlying domain. Later on, this model can be finetuned for a specific downstream task using significantly fewer annotations. With the advent of Vision Transformers (ViT) \cite{vit}, similar strategies such as Masked Image Modeling have been developed for computer vision~\cite{beit, mae, simmim}, showing equally high benefit in complex computer vision tasks.

Despite the advancements in computer vision and natural language processing, the progress of artificial intelligence methods in the medical field has been slow due to the lack of a sufficient amount of annotated data for developing data-driven approaches~\cite{PSI_AVA}. While the largest endoscopic dataset, Cholec80~\cite{cholec80}, only contains 200k images, computer vision datasets can reach hundreds of millions of images~\cite{JFT-300M}. Additionally, medical downstream tasks requiring complex annotations, such as pixel-wise segmentations, often have less than 10k images. To overcome this challenge, pretrained models can be used to reduce the need for labelled data. However, even though endoscopic videos differ significantly from natural images, most works in surgical data science rely on models pretrained on natural images.

In this study, we want to investigate the gap between computer vision and medical community to understand whether and when an endoscopy domain pretraining makes sense. Towards this objective, our contributions are threefold:
\begin{enumerate}
    \item We compile the largest publicly available collection of unlabelled endoscopic data, Endo700k, consisting of more than 700,000 unannotated raw images.
    \item We introduce the first publicly available endoscopy pretrained vision transformer, EndoViT.
    \item We analyze, through comprehensive experiments and ablation studies, the effect of endoscopy pretraining on the downstream tasks.
\end{enumerate}

\section{Methodology}
\subsection{Dataset Preparation}
Self-supervised pretraining is a powerful technique for enhancing the performance of deep learning models. However, it requires a vast amount of data to be effective. To this end, we have created the largest publicly available collection of raw endoscopic data, Endo700k. Endo700k is formed by combining nine publicly available Minimally Invasive Surgery (MIS) datasets, comprising more than 700,000 unannotated images. An overview of the individual datasets is provided in Table \ref{table:Datasets}. Endo700k encompasses a diverse set of endoscopic procedures, both manual and/or robot-assisted, with several surgery types such as prostatectomy, cholecystectomy, gastrectomy, proctocolectomy, rectal resection, and sigmoid resection. Furthermore, multiple different surgical actions, anatomies, and many surgical instruments, which are present in different shapes and sizes, are included. 


\begin{table*}[t!]
  \caption{An overview of the individual datasets that build our Endo700k dataset. The first nine datasets (ESAD - Cholec80) represent a unique collection of roughly 744k raw endoscopic images. Cholec80 and its subvariant CholecT45 are additionally used for downstream tasks of Surgical Phase Recognition and Action Triplet Detection, respectively. \label{table:Datasets}}
  \centering
  \renewcommand{\arraystretch}{1.2}
  \resizebox{\textwidth}{!}{\begin{NiceTabular}{ Wc{0.4cm}  Wc{2.4cm}  Wc{3.2cm}  Wc{1.4cm}  Wc{1.8cm} }[hvlines-except-corners,
  code-before=
    \rectanglecolor{DatasetNameColumn_title_color}{1-1}{2-2}
    \rectanglecolor{OtherColumns_title_color}{1-3}{2-5}
    \rectanglecolor{DatasetNameColumn_color_light}{3-1}{4-2}
    \rectanglecolor{gray!15}{3-3}{4-5}
    \rectanglecolor{DatasetNameColumn_color}{5-1}{6-2}
    \rectanglecolor{gray!35}{5-3}{6-5}
    \rectanglecolor{DatasetNameColumn_color_light}{7-1}{8-2}
    \rectanglecolor{gray!15}{7-3}{8-5}
    \rectanglecolor{DatasetNameColumn_color}{9-1}{10-2}
    \rectanglecolor{gray!35}{9-3}{10-5}
    \rectanglecolor{DatasetNameColumn_color_light}{11-1}{14-2}
    \rectanglecolor{gray!15}{11-3}{14-5}
    \rectanglecolor{DatasetNameColumn_color}{15-1}{20-2}
    \rectanglecolor{gray!35}{15-3}{20-5}
    \rectanglecolor{DatasetNameColumn_color_light}{21-1}{22-2}
    \rectanglecolor{gray!15}{21-3}{22-5}
    \rectanglecolor{DatasetNameColumn_color}{23-1}{26-2}
    \rectanglecolor{gray!35}{23-3}{26-5}
    \rectanglecolor{DatasetNameColumn_color_light}{27-1}{28-2}
    \rectanglecolor{gray!15}{27-3}{28-5}
    \rectanglecolor{DatasetNameColumn_color}{29-1}{30-2}
    \rectanglecolor{gray!35}{29-3}{30-5}
    ]
    %
    %
      \Block{2-1}{\makecell{\textcolor{white}{\textbf{\#}}}}
    & \Block{2-1}{\makecell{\textcolor{white}{\textbf{Dataset}}}}
    & \Block{2-1}{\makecell{\textcolor{white}{\textbf{Surgery Type}}}}
    & \Block{2-1}{\makecell{\textcolor{white}{\textbf{\# Surg.}}}}
    & \Block{2-1}{\makecell{\textcolor{white}{\textbf{\# Unique}}\\\textcolor{white}{\textbf{Images}}}} \\
                     &                                                                          &                                                                             &                           &                       \\
    %
    %
    
    \Block{2-1}{1}   & \Block{2-1}{\makecell{ESAD~\cite{esad}}}                                 & \Block{2-1}{\makecell{Robot-Assisted \\ Radical Prostatectomy}}             &  \Block{2-1}{4}           & \Block{2-1}{49544}    \\
                     &                                                                          &                                                                             &                           &                       \\    
                                      
    \Block{2-1}{2}   & \Block{2-1}{\makecell{LapGyn4 (v1.2)\\~\cite{lapgyn4}}}                  & \Block{2-1}{\makecell{Gynecologic \\ Laparoscopy}}                          &  \Block{2-1}{> 500}       & \Block{2-1}{38192}    \\
                     &                                                                          &                                                                             &                           &                       \\
                                      
    \Block{2-1}{3}   & \Block{2-1}{\makecell{Surgical\\Actions160~\cite{SurgicalActions160}}}   & \Block{2-1}{\makecell{Gynecologic \\ Laparoscopy}}                          &  \Block{2-1}{59}          & \Block{2-1}{761}      \\
                     &                                                                          &                                                                             &                           &                       \\

    \Block{2-1}{4}   & \Block{2-1}{\makecell{GLENDA (v1.0)\\~\cite{GLENDA}}}                    & \Block{2-1}{\makecell{Gynecologic \\ Laparoscopy}}                          &  \Block{2-1}{> 400}       & \Block{2-1}{1083}     \\
                     &                                                                          &                                                                             &                           &                       \\
                                      
    \Block{4-1}{5}   & \Block{4-1}{\makecell{hSDB -\\instrument~\cite{hsdb}}}                   & \Block{2-1}{\makecell{Laparoscopic \\ Cholecystectomy}}                     &  \Block{2-1}{24}          & \Block{4-1}{35576}    \\
                     &                                                                          &                                                                             &                           &                       \\
                     &                                                                          & \Block{2-1}{\makecell{Robotic \\ Gastrectomy}}                              &  \Block{2-1}{24}          &                       \\
                     &                                                                          &                                                                             &                           &                       \\

    \Block{6-1}{6}   & \Block{6-1}{\makecell{HeiCo~\cite{heico}}}                               & \Block{2-1}{\makecell{Laparoscopic \\ Proctocolectomy}}                     &  \Block{2-1}{10}          & \Block{6-1}{347257}   \\
                     &                                                                          &                                                                             &                           &                       \\
                     &                                                                          & \Block{2-1}{\makecell{Laparoscopic \\ Rectal Resection}}                    &  \Block{2-1}{10}          &                       \\
                     &                                                                          &                                                                             &                           &                       \\
                     &                                                                          & \Block{2-1}{\makecell{Laparoscopic \\ Sigmoid Resection}}                   &  \Block{2-1}{10}          &                       \\
                     &                                                                          &                                                                             &                           &                       \\

    \Block{2-1}{7}   & \Block{2-1}{\makecell{PSI-AVA~\cite{PSI_AVA}}}                           & \Block{2-1}{\makecell{Robot-Assisted \\ Radical Prostatectomy}}             &  \Block{2-1}{8}           & \Block{2-1}{73618}    \\
                     &                                                                          &                                                                             &                           &                       \\

    \Block{4-1}{8}   & \Block{4-1}{\makecell{DSAD~\cite{DSAD}}}                                 & \Block{2-1}{\makecell{Robot-Assisted \\ Rectal Resection}}                  &  \Block{4-1}{32}          & \Block{4-1}{13195}    \\
                     &                                                                          &                                                                             &                           &                       \\
                     &                                                                          & \Block{2-1}{\makecell{Robot-Assisted \\ Rectal Extirpation}}                &                           &                       \\
                     &                                                                          &                                                                             &                           &                       \\

    \Block{2-1}{9}   & \Block{2-1}{\makecell{Cholec80~\cite{cholec80}}}                         & \Block{2-1}{\makecell{Laparoscopic \\ Cholecystectomy}}                     &  \Block{2-1}{80}          & \Block{2-1}{184498}   \\
                     &                                                                          &                                                                             &                           &                       \\

    \Block{2-1}{10}  & \Block{2-1}{\makecell{CholecT45~\cite{rendezvous}}}                      & \Block{2-1}{\makecell{Laparoscopic \\ Cholecystectomy}}                     &  \Block{2-1}{45}          & \Block{2-1}{0}        \\
                     &                                                                          &                                                                             &                           &                       \\
 
  \end{NiceTabular}}
\end{table*}

\subsubsection{Dataset Collection:}The downstream evaluation experiments are conducted on the Cholec80 dataset~\cite{cholec80} and its subvariant CholecT45~\cite{rendezvous}. To eliminate any potential data leakage, we do not include any images that appear in their validation or test sets. Furthermore, all synthetic images are excluded. Outside the previously mentioned exceptions, we use all of the images from the nine datasets. For consistency, we always downsample to 1 FPS.

\subsection{Model Pretraining} 
\begin{figure}
\includegraphics[width=\textwidth]{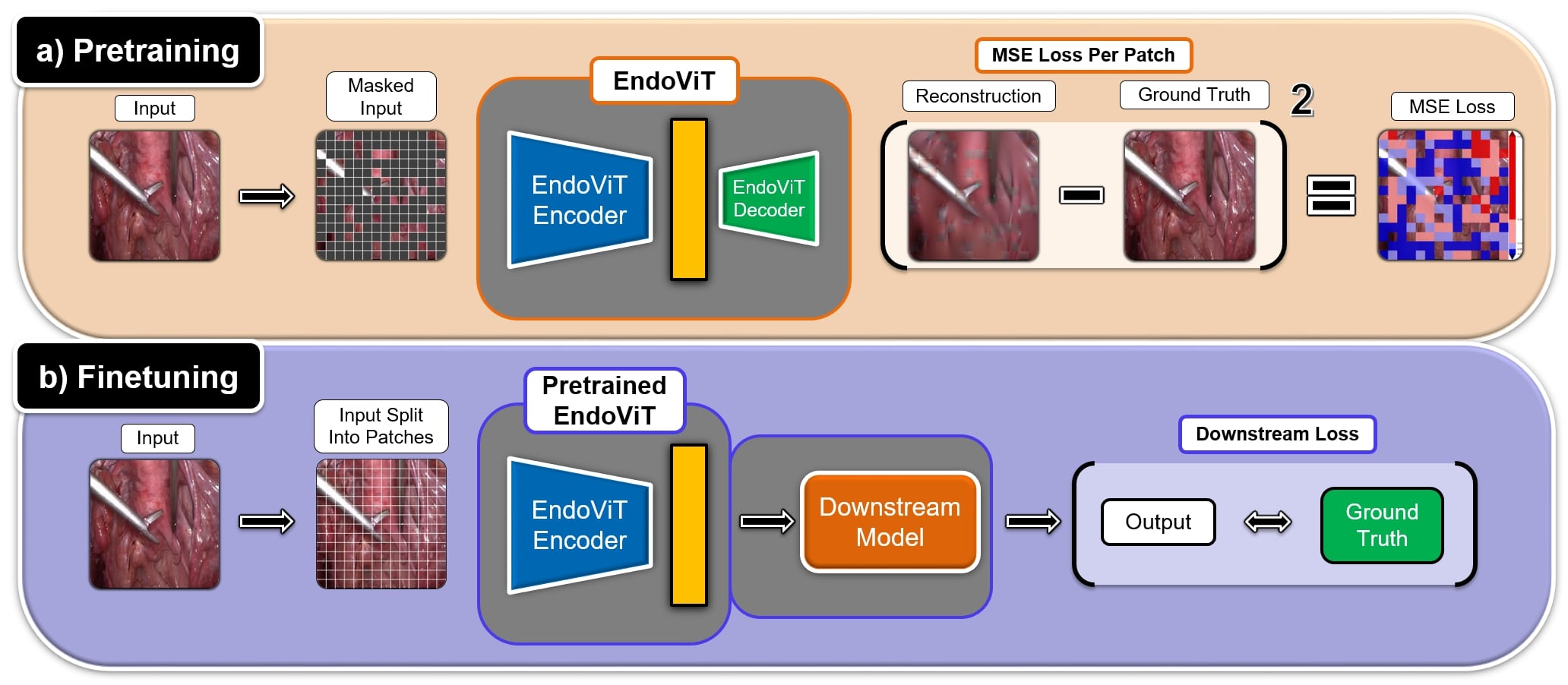}
\caption{The use of EndoViT is split into two parts: Pretraining and Finetuning. Our EndoViT model is first pretrained using the Masked Image Modeling strategy (a). An input image is split into non\hyp{}overlapping patches and a large proportion of them is masked out. The network then reconstructs the missing input in pixel space. Finally, per patch MSE loss is calculated on the missing patches only. By training the network to reconstruct the missing input, we obtain a model with a general understanding of endoscopic procedures. Later, EndoViT encoder can be used as a powerful feature extraction backbone on downstream tasks (b). When finetuning, an input image is still split into patches. However, to encode the full image, no masking is applied.
}
\label{fig:EndoViT_pretraining}
\end{figure}

We use Endo700k to pretrain EndoViT, our large-scale vision transformer based \cite{vit} feature extractor. A brief overview of the pretraining procedure can be seen in Fig. \ref{fig:EndoViT_pretraining}.

To this end, we closely follow the approach of MAE \cite{mae}, and employ the Masked Image Modeling strategy. The input image is first split into non\hyp{}overlapping patches. Afterward, a large proportion of them is masked out. By training the network to reconstruct the missing input, we produce a model with a general understanding of endoscopic procedures. The encoder of the pretrained model can then be used as a feature extraction backbone in the downstream tasks. We tailor the MAE approach for the endoscopic setting with three modifications:

\noindent \textbf{Layer Wise Learning Rate Decay:} The MAE encoder and decoder consist of several layers. We scale down the learning rate of each layer such that the layers closer to the latent space have larger learning rates, while those closer to the ends of the model have lower learning rates.

\noindent \textbf{Stochastic Weight Averaging (SWA)~\cite{swa}:} During the last 5 pretraining epochs, we average the models' weights at each validation step.

\noindent \textbf{Frequent Evalation:} The evaluation is performed 6 times per epoch, and the best SWA model is saved.

\subsubsection{Implementation Details:} We follow most of the practices of \cite{mae, msn}. During pretraining only simple image augmentations are applied, including random resized crop and random horizontal flips. We use AdamW optimizer \cite{adamw} with a learning rate of 1.5e-3 and batch size of 256. We pretrain for a total of 15 epochs. The training starts with 3 linear warmup epochs, continues according to the cosine scheduler until epoch 10, and ends with a constant learning rate applied while SWA is on. We use layer-wise learning rate decay of 0.65. Finally, Mean Squared Error (MSE) is used as the reconstruction loss. We pretrain two different models, one for each of the downstream tasks. Both are pretrained on Endo700k, however, the pretraining datasets are slightly different, obtained by removing validation and test datasets of CholecT45 and Cholec80, respectively. All models have been implemented in Pytorch 1.13.0 and trained on 1 Nvidia a40 GPU.

\subsection{Downstream Tasks}
After pretraining our feature extractor backbones, we evaluate their performance on two downstream tasks, namely Surgical Phase Recognition and Action Triplet Detection.

\subsubsection{Action Triplet Detection:}
In the task of Action Triplet Detection, the objective is to detect surgical instruments, actions and tissues, as well as their interactions. We build a straightforward model consisting of a feature extraction backbone and a linear head. Although the authors of CholecT45 propose a cross-validation setup, we decided to test only on fold 5. We do so for two reasons. First, testing only on one fold allows us to have CholecT45 images in the pretraining. Otherwise, all of the data would be considered a part of the test set and would have to be excluded. Secondly, CholecT45 is a highly imbalanced dataset. Positive classes make up only 2\% of all examples, and their distribution is also highly imbalanced. 15 out of 100 classes account for roughly 90\% of all positive instances. Choosing fold 5 as the test set gives us the most balanced distribution of classes across train, val, and test splits. While most works such as \cite{rendezvous, actiontripnet} utilize Binary Cross Entropy loss, we empirically find that Focal Loss brings significant improvement and therefore use it in all our experiments.

\subsubsection{Surgical Phase Recognition:}
In the task of Surgical Phase Recognition, the objective is to detect different phases of a surgical procedure based on the surgical video stream. For this task, we choose TeCNO~\cite{tecno}, a well-known benchmark model with publicly available code. TeCNO is a two-step surgical phase recognition method. In the first step, a single-frame ResNet50 model is trained to predict surgical phases. In the second step, a Multi-Stage Temporal Convolutional Network (MS-TCN) refines the extracted features using temporal context. This two-stage approach allows the MS-TCN to improve the predictions of any feature extractor regardless of the chosen architecture. In our experiments, we replace the Resnet50 backbone with a ViT model and otherwise stick to the training and evaluation setup of TeCNO.

\section{Experiments}
We compare our EndoViT with its ImageNet pretrained counterpart and most commonly used CNN architectures (ResNet50/ResNet18) in two different scenarios. We first train the models on the full downstream dataset. Afterward, we evaluate their few-shot learning performance using only 2, 4, or 8 training videos. We report the mean and standard deviation of the corresponding metrics across 3 runs for each network in each setting.

\subsection{Action Triplet Detection Results}

\subsubsection{Full Dataset:}
We report the results in Table \ref{table:Results_ATD_FDR}. The reported metric is mean Average Precision (mAP) proposed by the authors of CholecT45 dataset~\cite{rendezvous}. The results show that EndoViT outperforms both CNN architectures and its ImageNet pretrained counterpart by roughly 8\% and 2\%, respectively, empirically showcasing the value of using endoscopy-based models. Furthermore, from the performance of randomly initialized ViT model, it can be seen that pretraining is essential for vision transformers.

\begin{table}[h!]
    \centering
    \caption{Action Triplet Detection Full Dataset Results (mean Average Precision)\label{table:Results_ATD_FDR}}

    \resizebox{\textwidth}{!}{\begin{NiceTabular}{*{5}{c}}[hvlines-except-corners,
    code-before=
    \cellcolor{ResNet_color}{1-1, 1-2}
    \cellcolor{MAE_NoPretraining_color}{1-3}
    \cellcolor{MAE_ImageNet_color}{1-4}
    \cellcolor{EndoViT_color}{1-5}
    \rectanglecolor{gray!15}{2-1}{2-5}]
    \textcolor{white}{\textbf{ResNet18}}        & \textcolor{white}{\textbf{ResNet50}}        & \textcolor{white}{\textbf{ViT NoPretraining}}        & \textcolor{white}{\textbf{ViT ImageNet}}        & \textcolor{white}{\textbf{EndoViT}}        \\
    21.72\% ± 1.17\%                            & 22.13\% ± 1.37\%                            & 13.93\% ± 0.43\%                                     & 27.84\% ± 0.39\%                                & \textbf{30.17\% ± 0.01\%}                  \\
    \end{NiceTabular}}
\end{table}

\subsubsection{Few-Shot Experiments:}
In Table \ref{table:Results_ATD_FSE}, we report few-shot learning  experiment results by training only on 2, 4, or 8 videos, respectively. We keep the validation and test datasets the same. We observe the same trends in our few-shot learning experiments. EndoViT outperforms ResNet50 by rougly 5\% - 6.5\% and ImageNet model by roghly 2\% - 5\%.

\begin{table}[h!]
    \centering
    \caption{Action Triplet Detection Few-Shot Results (mean Average Precision)\label{table:Results_ATD_FSE}}

    \resizebox{\textwidth}{!}{\begin{NiceTabular}{*{4}{c}}[hvlines-except-corners,
    code-before=
    \cellcolor{ResNet_color}{1-2}
    \cellcolor{MAE_ImageNet_color}{1-3}
    \cellcolor{EndoViT_color}{1-4}
    \cellcolor{2VideosOnly_color}{2-1}
    \cellcolor{4VideosOnly_color}{3-1}
    \cellcolor{8VideosOnly_color}{4-1}
    \rectanglecolor{gray!15}{2-2}{2-5}
    \rectanglecolor{gray!35}{3-2}{3-5}
    \rectanglecolor{gray!15}{4-2}{4-5}]
                     & \textcolor{white}{\textbf{ResNet50}}        & \textcolor{white}{\textbf{ViT ImageNet}}        & \textcolor{white}{\textbf{EndoViT}}        \\
    2 Videos Only    & 10.88\% ± 0.50\%                            & 12.22\% ± 1.78\%                                & \textbf{17.59\% ± 2.94\%}                  \\
    4 Videos Only    & 12.37\% ± 1.78\%                            & 14.27\% ± 1.73\%                                & \textbf{18.52\% ± 2.28\%}                  \\
    8 Videos Only    & 17.01\% ± 1.75\%                            & 19.71\% ± 0.61\%                                & \textbf{21.91\% ± 0.12\%}                  \\
    \end{NiceTabular}}
    
\end{table}

\subsection{Surgical Phase Recognition Results}

\subsubsection{Full Dataset:}
We report the results in Table~\ref{table:Results_SPR_FDR}. The reported metric is Phase Accuracy averaged over all testing videos. We report the performance of all models after both stages of TeCNO training. For reference purposes, we note that the reported performance of ResNet50 backbone in TeCNO \cite{tecno} is 88.56\% ± 0.27\%. Once again, the CNN architecture was outperformed by the pretrained vision transformers. However, the difference between EndoViT and ImageNet pretrained backbone is negligible in both stages. We believe there are two causes. One, the semantic understanding induced by reconstructing image patches isn't capable of capturing long-term relationships required to accurately discriminate between different surgical phases. And two, the training set used in Cholec80 is large (approx. 90k images), making it easier to overcome the pretraining differences.

\begin{table}[h!]
    \centering
    \caption{Surgical Phase Recognition Full Dataset Results (mean Phase Accuracy)\label{table:Results_SPR_FDR}}

    \resizebox{\textwidth}{!}{\begin{NiceTabular}{*{5}{c}}[hvlines-except-corners,
    code-before=
    \cellcolor{ResNet_color}{1-2}
    \cellcolor{MAE_NoPretraining_color}{1-3}
    \cellcolor{MAE_ImageNet_color}{1-4}
    \cellcolor{EndoViT_color}{1-5}
    \cellcolor{Stage1_color}{2-1}
    \cellcolor{Stage2_color}{3-1}
    \rectanglecolor{gray!15}{2-2}{3-5}]
               & \textcolor{white}{\textbf{ResNet50}}        & \textcolor{white}{\textbf{ViT NoPretraining}}        & \textcolor{white}{\textbf{ViT ImageNet}}        & \textcolor{white}{\textbf{EndoViT}}        \\
    Stage 1    & 79.84\% ± 0.30\%                            & 59.21\% ± 0.36\%                                     & \textbf{82.94\% ± 0.69\%}                       & 82.60\% ± 1.26\%                           \\
    Stage 2    & 87.84\% ± 0.58\%                            & 73.42\% ± 0.70\%                                     & \textbf{89.56\% ± 0.65\%}                       & 89.37\% ± 0.95\%                           \\
    \end{NiceTabular}}
    
\end{table}

\subsubsection{Few-Shot Experiments:}
In Table \ref{table:Results_SPR_FSE}, we report few-shot learning  experiment results by training only on 2, 4 or 8 videos, respectively. We keep the validation and test datasets the same. ResNet50 showcases a significant decrease in performance. When training on 2 videos only, EndoViT outperforms its ImageNet counterpart in both stages. For 4 and 8 videos, we observe comparable performance.

\begin{table}[h!]
    \centering
    \caption{Surgical Phase Recognition Few-Shot Results (mean Phase Accuracy)\label{table:Results_SPR_FSE}}

    \resizebox{\textwidth}{!}{\begin{NiceTabular}{*{4}{c}}[hvlines-except-corners,
    code-before=
    \cellcolor{Stage1_color}{1-1}
    \cellcolor{ResNet_color}{1-2}
    \cellcolor{MAE_ImageNet_color}{1-3}
    \cellcolor{EndoViT_color}{1-4}
    \rectanglecolor{Stage1_lighter_color}{2-1}{2-1}
    \rectanglecolor{gray!15}{2-2}{2-4}
    \rectanglecolor{Stage1_light_color}{3-1}{3-1}
    \rectanglecolor{gray!35}{3-2}{3-4}
    \rectanglecolor{Stage1_lighter_color}{4-1}{4-1}
    \rectanglecolor{gray!15}{4-2}{4-4}
    \cellcolor{Stage2_color}{5-1}
    \cellcolor{ResNet_color}{5-2}
    \cellcolor{MAE_ImageNet_color}{5-3}
    \cellcolor{EndoViT_color}{5-4}
    \rectanglecolor{Stage2_lighter_color}{6-1}{6-1}
    \rectanglecolor{gray!15}{6-2}{6-4}
    \rectanglecolor{Stage2_light_color}{7-1}{7-1}
    \rectanglecolor{gray!35}{7-2}{7-4}
    \rectanglecolor{Stage2_lighter_color}{8-1}{8-1}
    \rectanglecolor{gray!15}{8-2}{8-4}]
    \textbf{Stage 1}        & \textcolor{white}{\textbf{ResNet50}}        & \textcolor{white}{\textbf{ViT ImageNet}}        & \textcolor{white}{\textbf{EndoViT}}        \\
    2 Videos Only           & 47.51\% ± 1.33\%                            & 63.59\% ± 1.07\%                                & \textbf{67.04\% ± 2.92\%}                  \\
    4 Videos Only           & 57.80\% ± 2.67\%                            & 67.72\% ± 0.90\%                                & \textbf{71.80\% ± 0.49\%}                  \\
    8 Videos Only           & 63.71\% ± 1.48\%                            & \textbf{75.50\% ± 0.32\%}                       & 75.30\% ± 1.83\%                           \\
    \textbf{Stage 2}        & \textcolor{white}{\textbf{ResNet50}}        & \textcolor{white}{\textbf{ViT ImageNet}}        & \textcolor{white}{\textbf{EndoViT}}        \\
    2 Videos Only           & 68.23\% ± 1.10\%                            & 77.05\% ± 1.71\%                                & \textbf{78.89\% ± 1.26\%}                  \\
    4 Videos Only           & 74.50\% ± 1.76\%                            & 80.00\% ± 0.62\%                                & \textbf{80.28\% ± 0.71\%}                  \\
    8 Videos Only           & 77.43\% ± 1.68\%                            & 84.10\% ± 0.38\%                                & \textbf{84.68\% ± 1.25\%}                   \\
    \end{NiceTabular}}
    
\end{table}

\section{Conclusion}
To summarise our results, in this work, we have observed the benefits of our EndoViT model, pretrained specifically for the endoscopy domain. While in all cases, EndoVit performed equally or better than the ImageNet pretrained model, the improvements were more pronounced in more complex tasks, such as Action Triplet Recognition. On the other hand, our results lead us to believe that simpler tasks and larger amounts of available finetuning data, make pretraining less impactful. Moreover, our pretraining implementation, which reconstructs image patches in pixel space, captures general information about the objects and scenes it has seen. However, it can't capture complex relationships, e.g. temporal cues between frames required for successful phase recognition. Despite these limitations, EndoViT is an excellent alternative to the ImageNet pretrained models and hopefully an introduction to future research into the pretraining methods tailored to medical computer vision.


\bibliographystyle{splncs04}
\bibliography{bibliography}

\begin{thebibliography}{10}
\providecommand{\url}[1]{\texttt{#1}}
\providecommand{\urlprefix}{URL }
\providecommand{\doi}[1]{https://doi.org/#1}

\bibitem{msn}
Assran, M., Caron, M., Misra, I., Bojanowski, P., Bordes, F., Vincent, P.,
  Joulin, A., Rabbat, M., Ballas, N.: Masked siamese networks for
  label-efficient learning. In: Computer Vision--ECCV 2022: 17th European
  Conference, Tel Aviv, Israel, October 23--27, 2022, Proceedings, Part XXXI.
  pp. 456--473. Springer (2022)

\bibitem{beit}
Bao, H., Dong, L., Piao, S., Wei, F.: {BEiT}: {BERT} pre-training of image
  transformers. In: International Conference on Learning Representations
  (2022), \url{https://openreview.net/forum?id=p-BhZSz59o4}

\bibitem{esad}
Bawa, V.S., Singh, G., KapingA, F., Skarga-Bandurova, I., Oleari, E., Leporini,
  A., Landolfo, C., Zhao, P., Xiang, X., Luo, G., et~al.: The saras endoscopic
  surgeon action detection (esad) dataset: challenges and methods. arXiv
  preprint arXiv:2104.03178  (2021)

\bibitem{DSAD}
Carstens, M., Rinner, F.M., Bodenstedt, S., Jenke, A.C., Weitz, J., Distler,
  M., Speidel, S., Kolbinger, F.R.: The dresden surgical anatomy dataset for
  abdominal organ segmentation in surgical data science. Scientific Data
  \textbf{10}(1), ~1--8 (2023)

\bibitem{chen2021transunet}
Chen, J., Lu, Y., Yu, Q., Luo, X., Adeli, E., Wang, Y., Lu, L., Yuille, A.,
  Zhou, Y.: Transunet: Transformers make strong encoders for medical image
  segmentation. arXiv preprint arXiv:2102.04306  (2021)

\bibitem{tecno}
Czempiel, T., Paschali, M., Keicher, M., Simson, W., Feussner, H., Kim, S.T.,
  Navab, N.: Tecno: Surgical phase recognition with multi-stage temporal
  convolutional networks. In: Medical Image Computing and Computer Assisted
  Intervention--MICCAI 2020: 23rd International Conference, Lima, Peru, October
  4--8, 2020, Proceedings, Part III 23. pp. 343--352. Springer (2020)

\bibitem{vit}
Dosovitskiy, A., Beyer, L., Kolesnikov, A., Weissenborn, D., Zhai, X.,
  Unterthiner, T., Dehghani, M., Minderer, M., Heigold, G., Gelly, S.,
  Uszkoreit, J., Houlsby, N.: An image is worth 16x16 words: Transformers for
  image recognition at scale. In: International Conference on Learning
  Representations (2021), \url{https://openreview.net/forum?id=YicbFdNTTy}

\bibitem{mae}
He, K., Chen, X., Xie, S., Li, Y., Doll{\'a}r, P., Girshick, R.: Masked
  autoencoders are scalable vision learners. In: Proceedings of the IEEE/CVF
  Conference on Computer Vision and Pattern Recognition. pp. 16000--16009
  (2022)

\bibitem{swa}
Izmailov, P., Wilson, A., Podoprikhin, D., Vetrov, D., Garipov, T.: Averaging
  weights leads to wider optima and better generalization. In: 34th Conference
  on Uncertainty in Artificial Intelligence 2018, UAI 2018. pp. 876--885 (2018)

\bibitem{jha2021kvasir}
Jha, D., Ali, S., Emanuelsen, K., Hicks, S.A., Thambawita, V., Garcia-Ceja, E.,
  Riegler, M.A., de~Lange, T., Schmidt, P.T., Johansen, H.D., et~al.:
  Kvasir-instrument: Diagnostic and therapeutic tool segmentation dataset in
  gastrointestinal endoscopy. In: MultiMedia Modeling: 27th International
  Conference, MMM 2021, Prague, Czech Republic, June 22--24, 2021, Proceedings,
  Part II 27. pp. 218--229. Springer (2021)

\bibitem{GLENDA}
Leibetseder, A., Kletz, S., Schoeffmann, K., Keckstein, S., Keckstein, J.:
  Glenda: gynecologic laparoscopy endometriosis dataset. In: MultiMedia
  Modeling: 26th International Conference, MMM 2020, Daejeon, South Korea,
  January 5--8, 2020, Proceedings, Part II 26. pp. 439--450. Springer (2020)

\bibitem{lapgyn4}
Leibetseder, A., Petscharnig, S., Primus, M.J., Kletz, S., M{\"u}nzer, B.,
  Schoeffmann, K., Keckstein, J.: Lapgyn4: a dataset for 4 automatic content
  analysis problems in the domain of laparoscopic gynecology. In: Proceedings
  of the 9th ACM multimedia systems conference. pp. 357--362 (2018)

\bibitem{adamw}
Loshchilov, I., Hutter, F.: Decoupled weight decay regularization. In:
  International Conference on Learning Representations (2019),
  \url{https://openreview.net/forum?id=Bkg6RiCqY7}

\bibitem{SurgicalDataScience}
Maier-Hein, L., Vedula, S.S., Speidel, S., Navab, N., Kikinis, R., Park, A.,
  Eisenmann, M., Feussner, H., Forestier, G., Giannarou, S., et~al.: Surgical
  data science for next-generation interventions. Nature Biomedical Engineering
   \textbf{1}(9),  691--696 (2017)

\bibitem{heico}
Maier-Hein, L., Wagner, M., Ross, T., Reinke, A., Bodenstedt, S., Full, P.M.,
  Hempe, H., Mindroc-Filimon, D., Scholz, P., Tran, T.N., et~al.: Heidelberg
  colorectal data set for surgical data science in the sensor operating room.
  Scientific data  \textbf{8}(1), ~101 (2021)

\bibitem{actiontripnet}
Nwoye, C.I., Gonzalez, C., Yu, T., Mascagni, P., Mutter, D., Marescaux, J.,
  Padoy, N.: Recognition of instrument-tissue interactions in endoscopic videos
  via action triplets. In: Medical Image Computing and Computer Assisted
  Intervention {\textendash} {MICCAI} 2020, pp. 364--374. Springer
  International Publishing (2020)

\bibitem{rendezvous}
Nwoye, C.I., Yu, T., Gonzalez, C., Seeliger, B., Mascagni, P., Mutter, D.,
  Marescaux, J., Padoy, N.: Rendezvous: Attention mechanisms for the
  recognition of surgical action triplets in endoscopic videos. Medical Image
  Analysis  \textbf{78},  102433 (2022)

\bibitem{SurgicalActions160}
Schoeffmann, K., Husslein, H., Kletz, S., Petscharnig, S., Muenzer, B., Beecks,
  C.: Video retrieval in laparoscopic video recordings with dynamic content
  descriptors. Multimedia Tools and Applications  \textbf{77},  16813--16832
  (2018)

\bibitem{JFT-300M}
Sun, C., Shrivastava, A., Singh, S., Gupta, A.: Revisiting unreasonable
  effectiveness of data in deep learning era. In: Proceedings of the IEEE
  international conference on computer vision. pp. 843--852 (2017)

\bibitem{twinanda2016endonet}
Twinanda, A.P., Shehata, S., Mutter, D., Marescaux, J., De~Mathelin, M., Padoy,
  N.: Endonet: a deep architecture for recognition tasks on laparoscopic
  videos. IEEE transactions on medical imaging  \textbf{36}(1),  86--97 (2016)

\bibitem{cholec80}
Twinanda, A.P., Shehata, S., Mutter, D., Marescaux, J., De~Mathelin, M., Padoy,
  N.: Endonet: a deep architecture for recognition tasks on laparoscopic
  videos. IEEE transactions on medical imaging  \textbf{36}(1),  86--97 (2016)

\bibitem{PSI_AVA}
Valderrama, N., Ruiz~Puentes, P., Hern{\'a}ndez, I., Ayobi, N., Verlyck, M.,
  Santander, J., Caicedo, J., Fern{\'a}ndez, N., Arbel{\'a}ez, P.: Towards
  holistic surgical scene understanding. In: Medical Image Computing and
  Computer Assisted Intervention--MICCAI 2022: 25th International Conference,
  Singapore, September 18--22, 2022, Proceedings, Part VII. pp. 442--452.
  Springer (2022)

\bibitem{vaswani2017attention}
Vaswani, A., Shazeer, N., Parmar, N., Uszkoreit, J., Jones, L., Gomez, A.N.,
  Kaiser, {\L}., Polosukhin, I.: Attention is all you need. Advances in neural
  information processing systems  \textbf{30} (2017)

\bibitem{simmim}
Xie, Z., Zhang, Z., Cao, Y., Lin, Y., Bao, J., Yao, Z., Dai, Q., Hu, H.:
  Simmim: A simple framework for masked image modeling. In: Proceedings of the
  IEEE/CVF Conference on Computer Vision and Pattern Recognition. pp.
  9653--9663 (2022)

\bibitem{hsdb}
Yoon, J., Lee, J., Heo, S., Yu, H., Lim, J., Song, C.H., Hong, S., Hong, S.,
  Park, B., Park, S., et~al.: hsdb-instrument: instrument localization database
  for laparoscopic and robotic surgeries. In: Medical Image Computing and
  Computer Assisted Intervention--MICCAI 2021: 24th International Conference,
  Strasbourg, France, September 27--October 1, 2021, Proceedings, Part IV 24.
  pp. 393--402. Springer (2021)

\end{thebibliography}


\begin{figure}
\includegraphics[width=\textwidth]{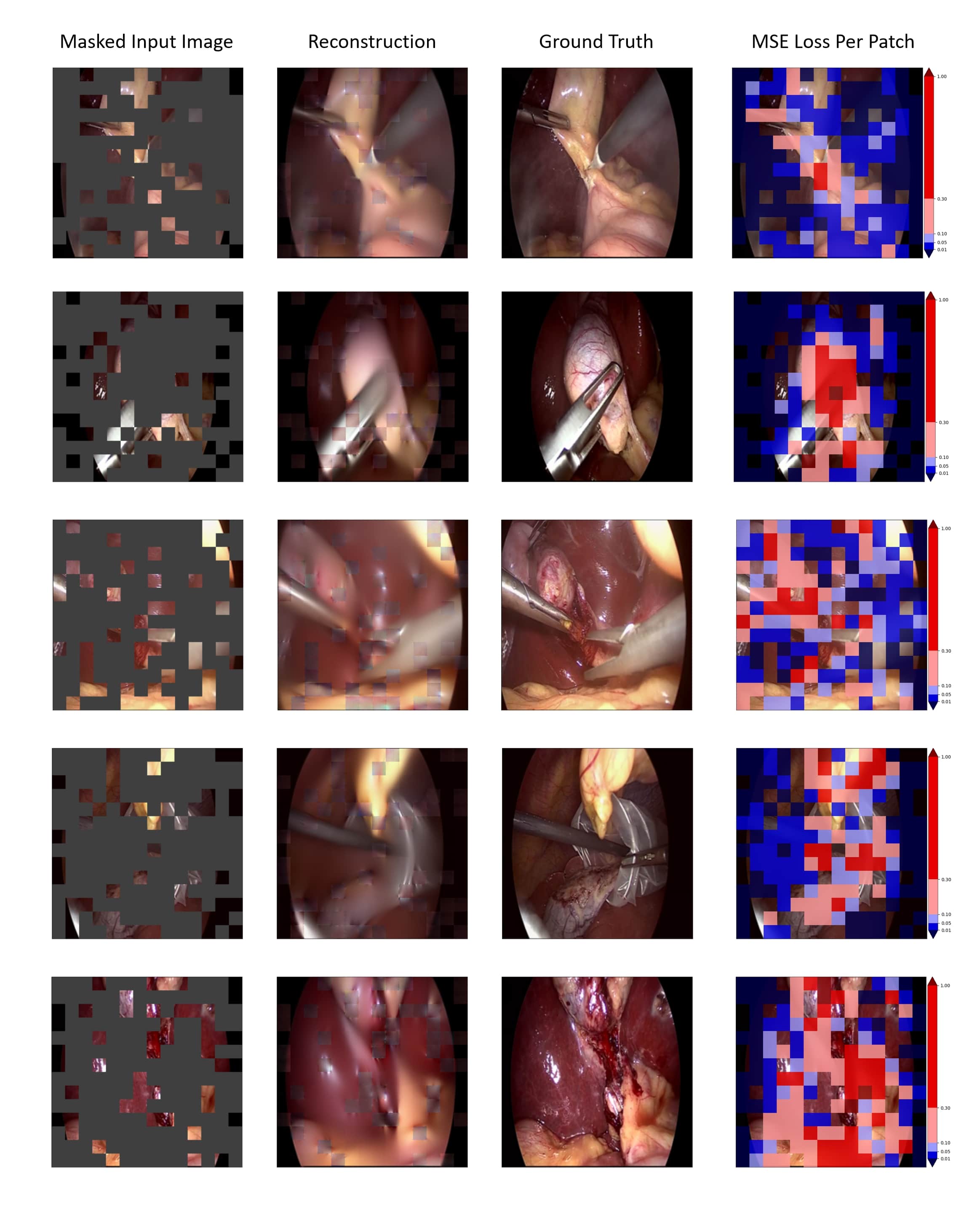}
\caption{Qualitative pretraining results: Random masking patterns applied to different test images. Every row represents one example. For each example, we show the masked input image (1st column), the model's reconstruction of the missing input (2nd column), the ground truth image (3rd column) and finally, per patch MSE loss which is calculated on the missing patches only (4th column). Red - higher loss, Blue - lower loss.} \label{Fig_supl2}
\end{figure}

\begin{figure}
\includegraphics[width=\textwidth]{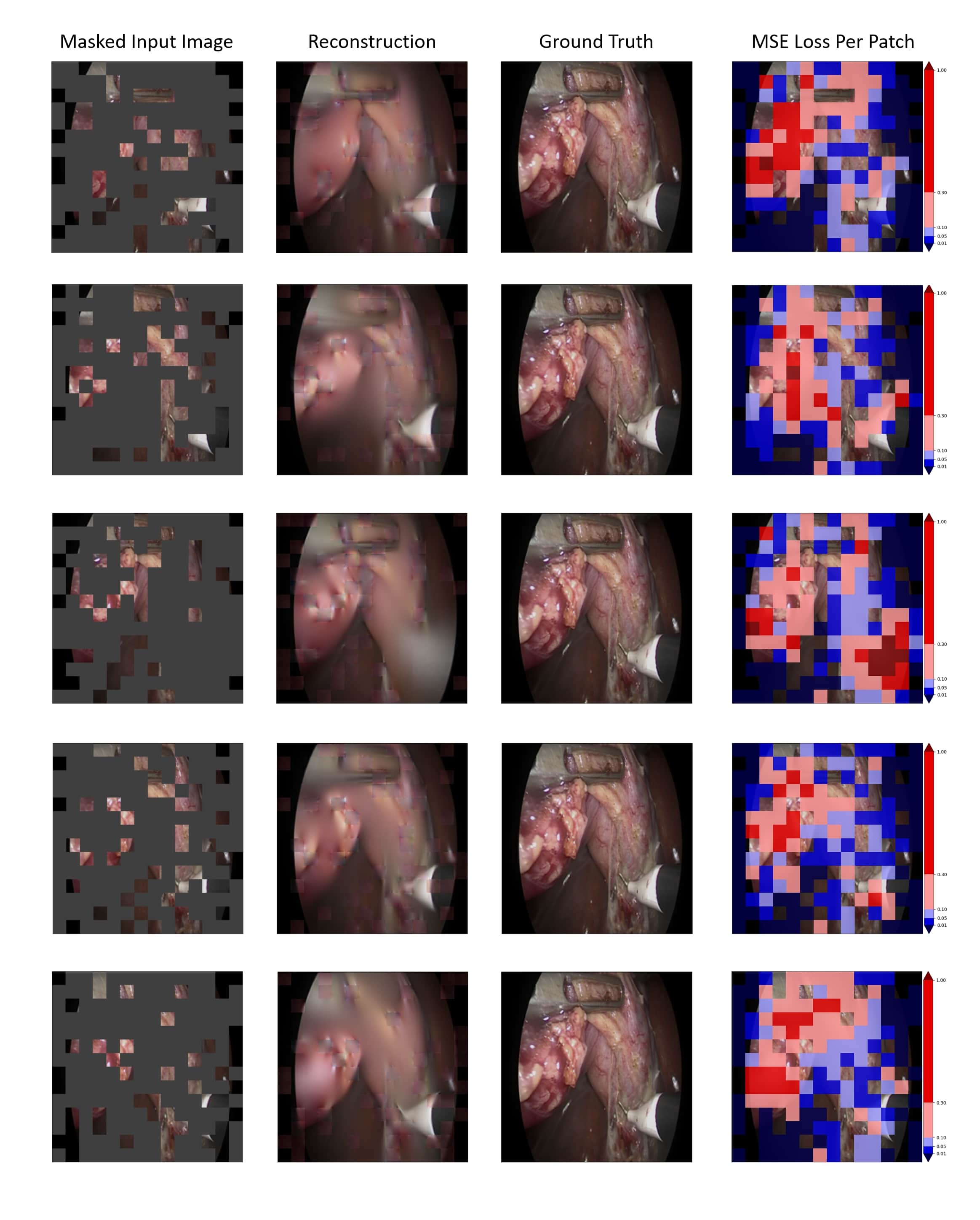}
\caption{Qualitative pretraining results: Different masking patterns applied to the same test image. Every row represents one pattern. For each example, we show the masked input image (1st column), the model's reconstruction of the missing input (2nd column), the ground truth image (3rd column) and finally, per patch MSE loss which is calculated on the missing patches only (4th column). Red - higher loss, Blue - lower loss.} \label{Fig_supl3}
\end{figure}


\end{document}